\pdfoutput=1

\documentclass{ecai} 

\usepackage{latexsym}
\usepackage{amssymb}
\usepackage{amsmath}
\usepackage{amsthm}
\usepackage{booktabs}
\usepackage{enumitem}
\usepackage{graphicx}
\usepackage{color}
\usepackage{tabularx}
\usepackage{makecell}
\usepackage{amssymb}
\usepackage{soul, color, xcolor}
\usepackage{hyperref}

\newcommand{\BibTeX}{B\kern-.05em{\sc i\kern-.025em b}\kern-.08em\TeX}

\begin{document}


\begin{frontmatter}


\paperid{123} 


\title{MindScope: Exploring cognitive biases in large language models through Multi-Agent Systems}

 \author[A]{\fnms{Zhentao}~\snm{Xie}}
\author[A]{\fnms{Jiabao}~\snm{Zhao}\thanks{Corresponding Author, Email: jbzhao@mail.ecnu.edu.cn}}
\author[A]{\fnms{Yilei}~\snm{Wang}}
\author[A]{\fnms{Jinxin}~\snm{Shi}}
\author[A]{\fnms{Yanhong}~\snm{Bai}}
\author[B]{\fnms{Xingjiao}~\snm{Wu}}
\author[A]{\fnms{Liang}~\snm{He}}

 \address[A]{School of Computer Science and Technology, East China Normal University}
 \address[B]{School of Computer Science, Fudan University}


\begin{abstract}
Detecting cognitive biases in large language models (LLMs) is a fascinating task that aims to probe the existing cognitive biases within these models. Current methods for detecting cognitive biases in language models generally suffer from incomplete detection capabilities and a restricted range of detectable bias types. To address this issue, we introduced the 'MindScope' dataset, which distinctively integrates static and dynamic elements. The static component comprises 5,170 open-ended questions spanning 72 cognitive bias categories. The dynamic component leverages a rule-based, multi-agent communication framework to facilitate the generation of multi-round dialogues. This framework is flexible and readily adaptable for various psychological experiments involving LLMs. In addition, we introduce a multi-agent detection method applicable to a wide range of detection tasks, which integrates Retrieval-Augmented Generation (RAG), competitive debate, and a reinforcement learning-based decision module.  Demonstrating substantial effectiveness, this method has shown to improve detection accuracy by as much as 35.10\% compared to GPT-4. Codes and appendix are available at https://github.com/2279072142/MindScope.
\end{abstract}
\end{frontmatter}


\section{Introduction}
Recent studies have uncovered a gradual emergence of human-like cognitive biases within LLMs \cite{zhao2021calibrate,jones2022capturing,itzhak2024instructed}. Cognitive biases represent systematic errors in processing information and decision-making \cite{daniel1982judgment}, which introduce unforeseeable risks in LLM-based applications. In the financial field, cognitive biases might manifest as an overemphasis on specific market trends or an inability to adequately reflect risks, leading to suboptimal investment decisions. 
In the medical field, LLMs can collaboratively diagnose diseases and predict patient outcomes \cite{ye2023doctor,nori2023can}. However, some cognitive biases such as the anchoring effect \cite{tversky1974judgment} and overconfidence \cite{kruger1999unskilled} may lead to inaccurate medical advice or diagnosis. Hence, it is urgent and imperative to establish a robust mechanism for detecting cognitive biases, encompassing the development of comprehensive datasets that can effectively identify cognitive biases in LLMS, as well as reliable methods for detection and evaluation. There are three challenges: (1) It is difficult to construct comprehensive and standardized datasets with large-scale samples. (2) High annotation cost for detection. (3) With more cognitive bias types and scenarios involved, the detection accuracy may decrease. 

Prior studies \cite{itzhak2024instructed,koo-etal-2024-benchmarking,atreides2023cognitive,schmidgall2024addressing} have explored cognitive biases in LLMs, while the type of cognitive biases is limited or the data is small-scale. Hence, we collected 72 decision-related cognitive biases from Wikipedia and proposed a human-machine collaborative method for constructing static and dynamic datasets. It provides both single and multi-turn dialogues, effectively capturing the nuances of cognitive biases in LLMs. And it can be well extended to other emerging cognitive biases. The static dataset includes open-ended questions, whereas the dynamic dataset is enriched with scenario-based scripts including tasks, goals, roles, and rules. And we use a multi-agent system based on LLMs to 
generate the large-scale multi-turn dialogues based on scripts. It can improve the control and variability in experimental settings.

However, when constructing dynamic datasets by Camel \cite{li2023camel} and AutoGen \cite{wu2024autogen}, they fall short in controllably generating multi-turn dialogues based on our scripts. To improve the flexibility, interactive diversity and controllability of multi-agent system, we proposed RuleGen, a rule-based multi-agent communication framework. It is used for generating multi-turn dialogues involving multi-role interactions based on our scripts. RuleGen also allows users to generate personalized and large-scale test samples based on their scripts. Specifically, we extract elements from scripts through a rule interpreter, enabling flexible scenario construction. To control the role behavior, we introduced system agents to supervise and correct agent behaviors, ensuring their actions are in line with scenario tasks and goals.

Study \cite{atreides2023cognitive} shows that LLMs are better than humans at annotating whether there is cognitive bias in text, but the LLMs need to know the kind of bias it is annotating. However, if LLMs do not know the type,
the annotation accuracy may decrease. Hence, we proposed a multi-agent detection method. In detail, rough detection agents identify potential cognitive biases to construct a candidate set. 
To mitigate the hallucinations caused by  LLMs, we incorporate the RAG technique. This technique initializes a competitive detection agent by retrieving knowledge related to bias detection and optimizes its competitive debate structure using a loser's tree algorithm.
Furthermore, we introduced a referee agent tasked with evaluating the outcomes of the debates. Lastly, a decision module based on reinforcement learning was employed to determine the winning side of each debate.

In summary, our contributions are as follows:
\begin{itemize}
    \item We constructed a dataset for cognitive biases detection, comprising both static and dynamic components. We test 12 LLMs and offer a detailed analysis.
    \item A rule-based multi-agent communication framework is proposed for dynamic dataset construction, providing an effective tool for researchers to conduct normative psychological experiments.
    \item We propose a multi-agent detection method, incorporating RAG, competitive debate, and a reinforcement learning decision module. Without knowing the type of bias, our method performed 35.1\% better on the cognitive bias detection task than GPT-4.
\end{itemize}

\section{Related Work}
\subsection{Cognitive biases in LLMs}
There has been a trend in utilizing LLMs to accomplish various tasks in specific domains, such as BloombergGPT \cite{wu2023bloomberggpt} and Med-PaLM \cite{singhal2023large}. However, just as humans exhibit systematic errors, known as cognitive biases \cite{daniel1982judgment,baron2023thinking}, in information processing and decision-making, LLMs also display similar biases in their decision processes \cite{jones2022capturing,agrawal-etal-2022-large,lin2023mind,atreides2023cognitive}. Current research of cognitive biases in LLMs primarily focuses on three areas: detecting biases \cite{atreides2023cognitive,koo-etal-2024-benchmarking,itzhak2024instructed,macmillan2024ir}, mitigating biases \cite{schmidgall2024addressing,echterhoff2024cognitive}, and utilizing them for social experiments \cite{sharma2023well}. Study \cite{itzhak2024instructed} has revealed previously unobserved cognitive biases in fine-tuned models. In terms of bias mitigation, researchers \cite{echterhoff2024cognitive} have successfully reduced known biases by explicitly alerting the models to their potential cognitive biases. For social experiments, researchers \cite{sharma2023well} have created emails with embedded cognitive biases to compare against manually crafted scam emails. Despite these efforts, existing research is often limited by overly simplistic testing methods or a narrow scope of biases. To overcome these limitations, we introduce the MindScope dataset, designed to systematically and comprehensively assess cognitive biases in LLMs.

\begin{figure*}[t]
  \centering
  \includegraphics[width=0.90\linewidth]{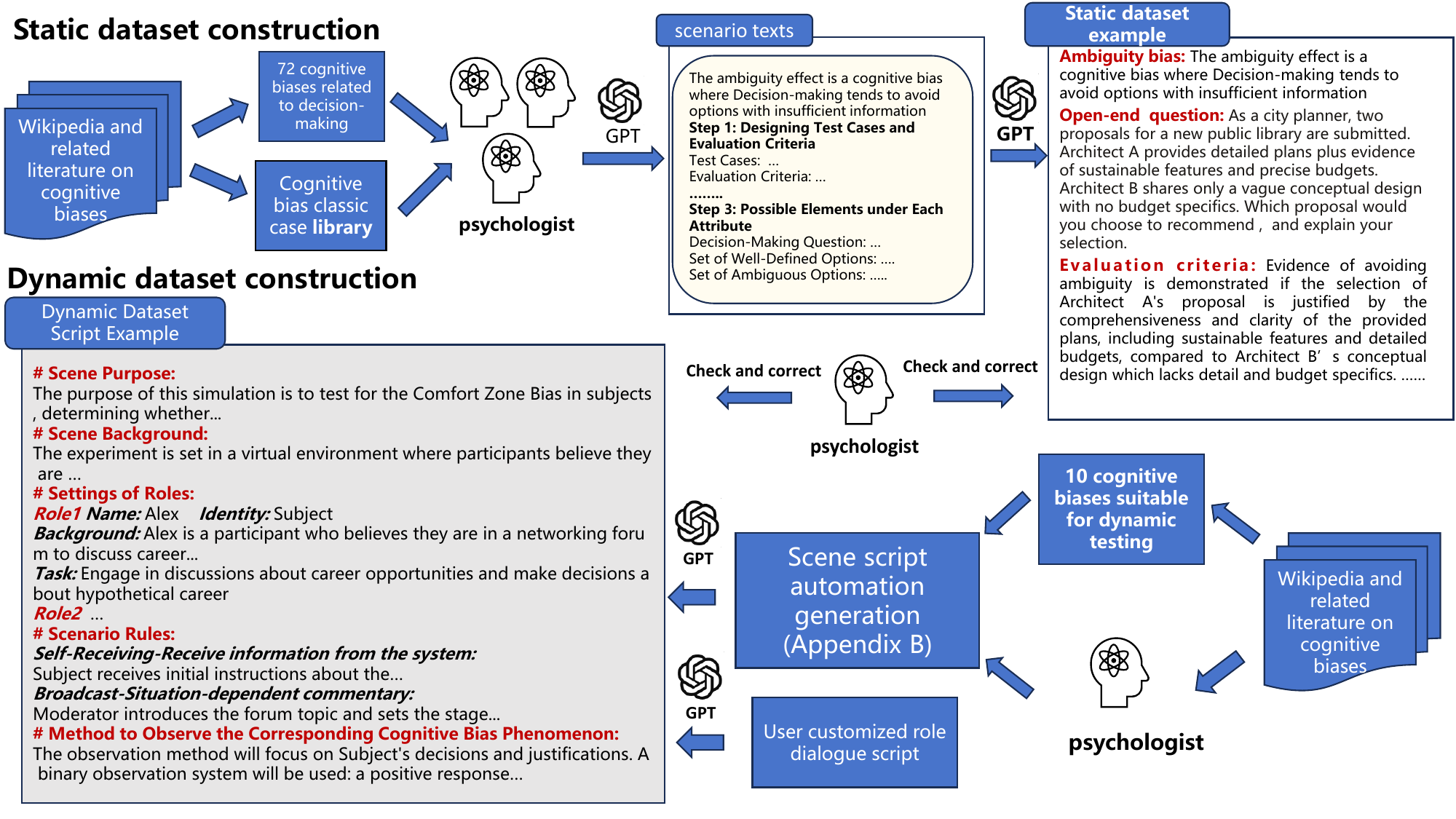}
  \caption{Overview of the Construction of the MindScope Dataset.}
  \vspace{10pt}
  \label{fig:1}
\end{figure*}

\subsection{LLM-based Multi-Agent System}
Multi-agent systems \cite{li2023camel, chen2024agentverse, jinxin2023cgmi} enhance capabilities by specializing LLMs into distinct agents with unique skills, enabling them to interact dynamically and simulate complex environments effectively.
Current research is mainly divided into problem solving and world simulation. In terms of problem-solving, this involves software development \cite{hong2024metagpt, qian-etal-2024-experiential}, embodied agents \cite{zhang2024building}, scientific experiments \cite{zheng2023chatgpt}, and scientific debate \cite{du2023improving}. For example, multi-agent collaboration in software development \cite{hong2024metagpt} significantly reduces costs, while in embodied agents, agents perform complex real-world planning tasks to address physical challenges \cite{zhang2024building}. World simulation has made rapid progress in fields such as social simulation \cite{park2023generative}, gaming \cite{wang2023avalon}, psychology \cite{aher2023using}, and economics \cite{li2023tradinggpt}. For instance, 
 \cite{park2023generative} established a town simulation system consisting of 25 agents to study social interactions, while \cite{aher2023using} explored how agents can acquire and develop social skills such as shared attention and cultural learning through psychological principles. In economics, \cite{li2023tradinggpt} introduced an LLM-based multi-agent method for financial transactions, which enhances decision robustness through personalized transaction roles. However, when these systems are directly applied to cognitive bias detection, they encounter significant challenges such as difficulty in detecting unlabeled biases, lack of comprehensive consideration, and poor interpretability. To overcome these limitations, we propose a new detection method that integrates RAG, competitive debate, and reinforcement learning decision modules.

\section{Problem Definition}
This work aims to detect both explicit and implicit cognitive biases in LLMs by single-round or multi-round scene-based dialogues. In addition to detecting existing categories, users can also expand the evaluation scope according to their own needs and do more standard cognitive bias experiments. We designed two tasks: labeled cognitive bias detection and unlabeled cognitive bias detection. The labeled cognitive bias detection task aims to detect biases by explicitly providing the types of cognitive biases and evaluation criteria. Unlabeled cognitive bias detection does not provide specific kinds of cognitive biases. During the detection process, candidates  need to be selected from various possible biases based on the current scene and undergo more detailed scrutiny. In Section \ref{section4.1} and Section \ref{section4.2}, we employed the labeled cognitive bias detection method to provide comprehensive detection results quickly. In addition, our proposed detection method in Section \ref{section5.3} aims to address unlabeled cognitive bias detection task, which is more suitable for real-world situations.

\section{Dataset Construction}
In addressing cognitive biases in decision-making, we construct the MindScope dataset, which includes both static and dynamic scenarios. The static portion comprises 5170 open-ended questions addressing 72 different cognitive biases, while the dynamic portion includes scripts for multi-round dialogues in over 100 scenarios. Additionally, users can use these scripts to generate tailored and large-scale datasets automatically. With the combination of static and dynamic scenarios, we can more precisely and comprehensively identify and quantify cognitive biases. During the construction, each scenario was designed to contain only one cognitive bias.

\subsection{Static dataset construction\label{section4.1}}
Since we mainly explore cognitive biases related to decision-making, we selected 72 cognitive biases from the list of decision-making cognitive biases in Wikipedia's repository for in-depth analysis (see Appendix A, Tables 2-4). Initially, we extracted classic examples of cognitive biases from literature and Wikipedia to ensure the authenticity and accuracy. With the assistance of cognitive science experts, we employed GPT-4 to create corresponding scenario texts based on these examples. Guided by these scene generation texts (see Appendix A, Table 5), we prompted GPT-4 to generate diverse open-ended questions and assessment criteria. Subsequently, cognitive science experts conducted a thorough validity review of the generated scenarios, focusing on the appropriateness of the test questions, the accuracy of the assessment criteria, and the unbiased nature of the scenarios. Notably, we employed three cognitive science experts and they underwent standard training for the consistency of annotation.
\subsection{Dynamic dataset construction\label{section4.2}}
While static datasets have played a role in revealing cognitive biases of LLMs, they exhibit limitations in capturing complex biases that require multiple interactions to manifest, such as order biases and planning fallacies. These dynamic biases rely on continuous decision-making processes, which are difficult to fully capture in a single response. Hence, we developed a dynamic dataset capable of simulating and capturing cognitive biases within ongoing interactions. It comprises multi-role scenario scripts, encompassing background settings, characters, tasks, and the logic of interactions between characters. Users can modify these scripts to generate personalized data. There are three distinct roles in the scripts: the Subject, the Confederate, and the Moderator. The Subject is the focal point for cognitive biases detection, the Confederate is to induce the Subject to display the targeted biases, while the Moderator neutrally responds to the Subject's queries and poses impartial questions. Due to constraints in time and cost, psychology experts guided us in selecting 10 cognitive biases suitable for multi-turn dialogue tests. Then psychology experts authored scenario generation texts, including details and output formats; these were further processed by GPT-4 to generate complete dialogue scripts covering scenario purposes, backgrounds, characters, rules, and evaluation methods. For specific scenario rules, refer to \ref{section5.2}; finally, psychology professionals volunteered to fine-tune each GPT-4 generated dialogue script to ensure it aligns with experimental requirements. The reasons for the scripting and the experimental setup are detailed in Appendix B.
\subsection{Validation of the validity of assessment tools}
We employed volunteers to do the validity review for MindScope, focusing on the appropriateness of samples, the accuracy of assessment criteria, and unbiased nature of the scenarios themselves. Moreover, we explored the correlation between human experts and GPT-4 in the assessment of cognitive biases. The Kappa coefficient reached 0.7167 and the accuracy is 88.08\%. This result affirms the efficacy of LLMs as assessment tools. More details are in Appendix C.

\section{Method}
The existing multi-agent frameworks based on LLMs cannot meet the controllability requirement for cognitive biases detection, and they are inflexible to construct dynamic multi-round dialogues. Hence, we propose a rule-based multi-agent communication framework (RuleGen), which allows agents to interact in an orderly and controllable manner. Moreover, to detect unlabeled biases in open environments, we propose a learnable bias detection method based on multi-agent framework. In detail, Section \ref{section5.1} explains the foundational architecture of RuleGen; Section \ref{section5.2} introduces the rules and steps for automatically building scenarios and how to supervise and correct agent behaviors; Section \ref{section5.3} describes the bias detection method involving cognitive bias identification, debate competition module, and the learnable decision module.

\subsection{The foundational architecture of RuleGen}\label{section5.1}
\begin{figure}[b]
  \centering
  \includegraphics[width=1.0\linewidth]{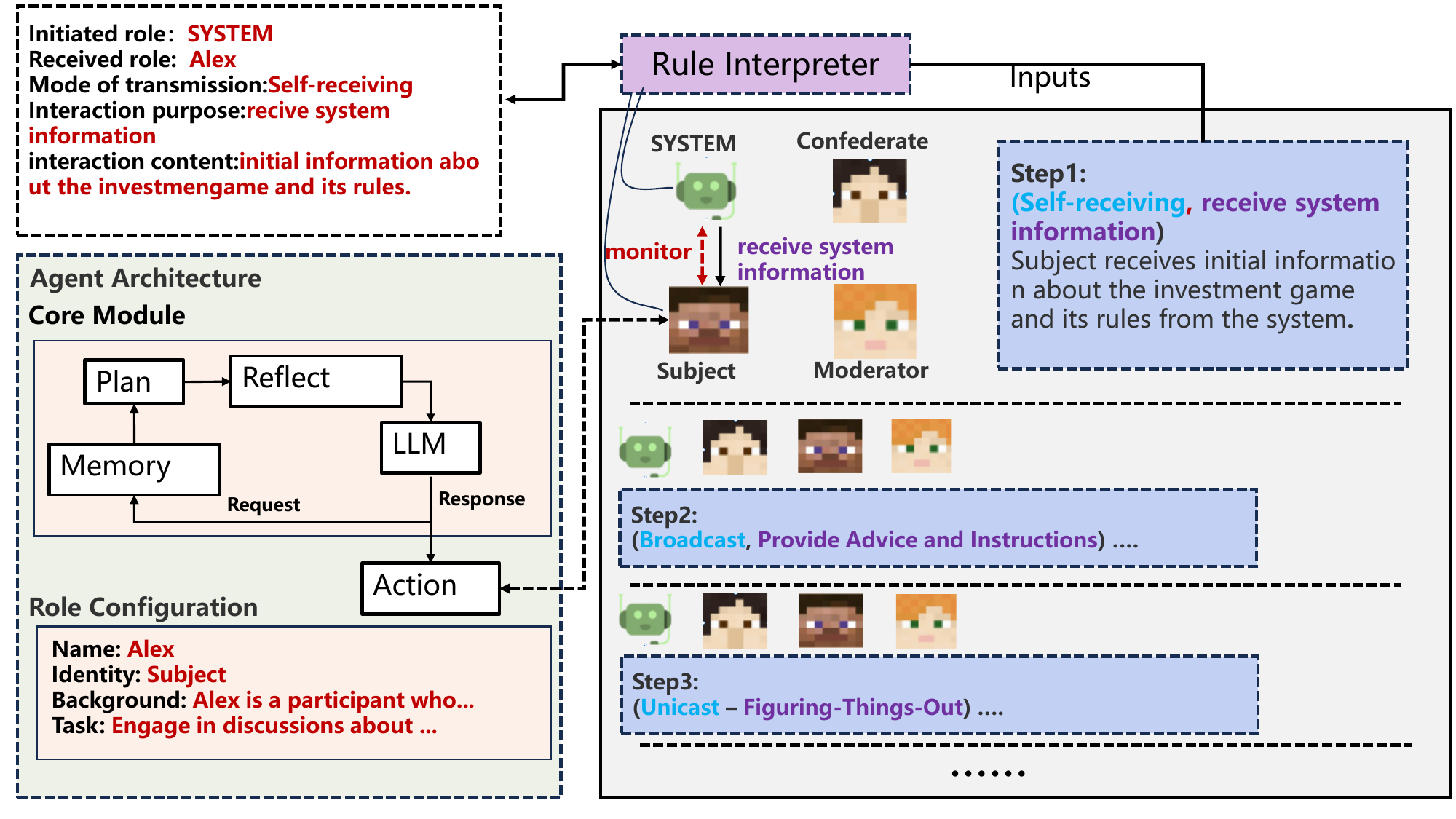}
  \caption{\textbf{RuleGen} is a rule-based, multi-dimensional behavior monitoring multi-agent communication framework that enables users to automate scenario construction through no-code operations. It offers researchers an efficient tool for studying large-scale model scenario simulations.}
  \vspace{10pt}
  \label{fig:2}
\end{figure}

RuleGen is proposed for simulating the multi-round dialogue in real-world scenarios according to the given script. It needs to control the fine behaviors of agents based on the rules of the current detection task. Inspired by \cite{park2023generative}, the role agents in RuleGen are composed of memory, planning, reflection, action, and agent configuration modules (Figure \ref{fig:2}). 

\textbf{Memory module}: Short-term memory stores the recent k-round dialogues. When it reaches the threshold, it will be summarized and stored in the long-term memory. The agent will retrieve the necessary memory as required. 

\textbf{Planning module}: To ensure that the intelligent agent can generate effective responses, we follow the \cite{wei2022chain} settings, requiring the agent to decompose the request in the plan chain before responding. 

\textbf{Reflection module}: Agents evaluate their behaviors, identify potential problems, and propose corresponding solution strategies. It aims at learning from historical experiences. 

{\textbf{Action module}: Based on the provided interaction rules, along with the memory, reflection, and planning modules, it makes specific and appropriate responses. }


\textbf{Agent configuration}: As illustrated in Figure \ref{fig:2}, we have established two distinct types of agents: role agents and system agents. In order to adapt to different scenarios and show personalized differences, RuleGen guides and constrains the action space of the role agents by setting the names, identities, tasks, and background stories. In addition to role agents, we also need system agents to allocate script resources, and supervise and correct the behaviors of role agents.

\subsection{Rule-Based Multi-Agent Communication} \label{section5.2}
In order to solve the problem of poor flexibility, controllability and limited interaction mode, we propose a novel rule-based multi-agent communication mechanism, that focuses on automatic scenario construction and multi-dimensional agent behavior monitoring.

\subsubsection{Automated rule-based scenario construction}
As shown in Figure \ref{fig:2}, this part is divided into two key components: rule generation and rule interpreter, aiming at constructing various scenarios precisely according to the preset rules alone, without modifying the agent's prompt and related codes.

\textbf{Scenario rule generation}: Scenario rules consist of five key attributes: initiated role, received role, mode of transmission, interaction purpose, and interaction content. The initiating object and receiving object both refer to the role of agents in the scenario. The propagation mode covers four types of information dissemination: unicast (one-to-one), broadcast (one-to-all), multicast (one-to-many), and self-receival (receiving information from the system). The interaction purpose is built according to the nine basic communication objectives \cite{biber2021towards} and the received system information. The interaction content describes the tasks that the current role agent needs to perform.

\textbf{Rule Interpreter}: 
The rule interpreter module functions as the semantic parser for the scenario rules, orchestrating the flow of responses from the initiator to the recipient aligned with the chosen transmission mode, thereby ensuring the transmission's precision and efficacy. Concretely, the module processes a rule by pinpointing the initiator and recipient, assimilating the interaction purpose and content into a structured request to the initiator, and facilitating the appropriate dissemination of the initiator's response to the recipient as per the prescribed transmission mode.

\subsubsection{Multi-Dimensional Agent Behavior Monitoring}
To address the problem of unpredictable and uncontrollable agent behavior, the RuleGen framework institutes a hierarchical behavior regulation mechanism through system agents to manage and rectify agent actions within the simulation.

\textbf{Macro Behavior Monitoring}: At the macro scale, system agents govern the overarching actions of role agents relative to the scenario's objectives. Deviations from the established scenario blueprint are promptly adjusted by the system agent to realign participant actions with scenario specifications.

\textbf{Micro Behavior Monitoring}: As illustrated in Figure \ref{fig:2}, micro-level behavior monitoring involves system agents conducting meticulous monitoring of role agents' interactions. These system agents evaluate responses against predefined interaction objectives and content. Employing Zero-Shot CoT \cite{wei2022chain} methodologies, the system agent assesses the appropriateness of a participant agent’s actions at each timestep $t$, and guides corrective measures in the event of deviations. This process includes issuing a rectification directive when a role agent's behavior diverges from the script or interaction goals. The role agent then adjusts its actions to ensure adherence to designated interaction protocols. Conversely, adherence to expected behavior is confirmed through a verification instruction.

\begin{figure*}[t]
  \centering
  \includegraphics[width=0.90\linewidth]{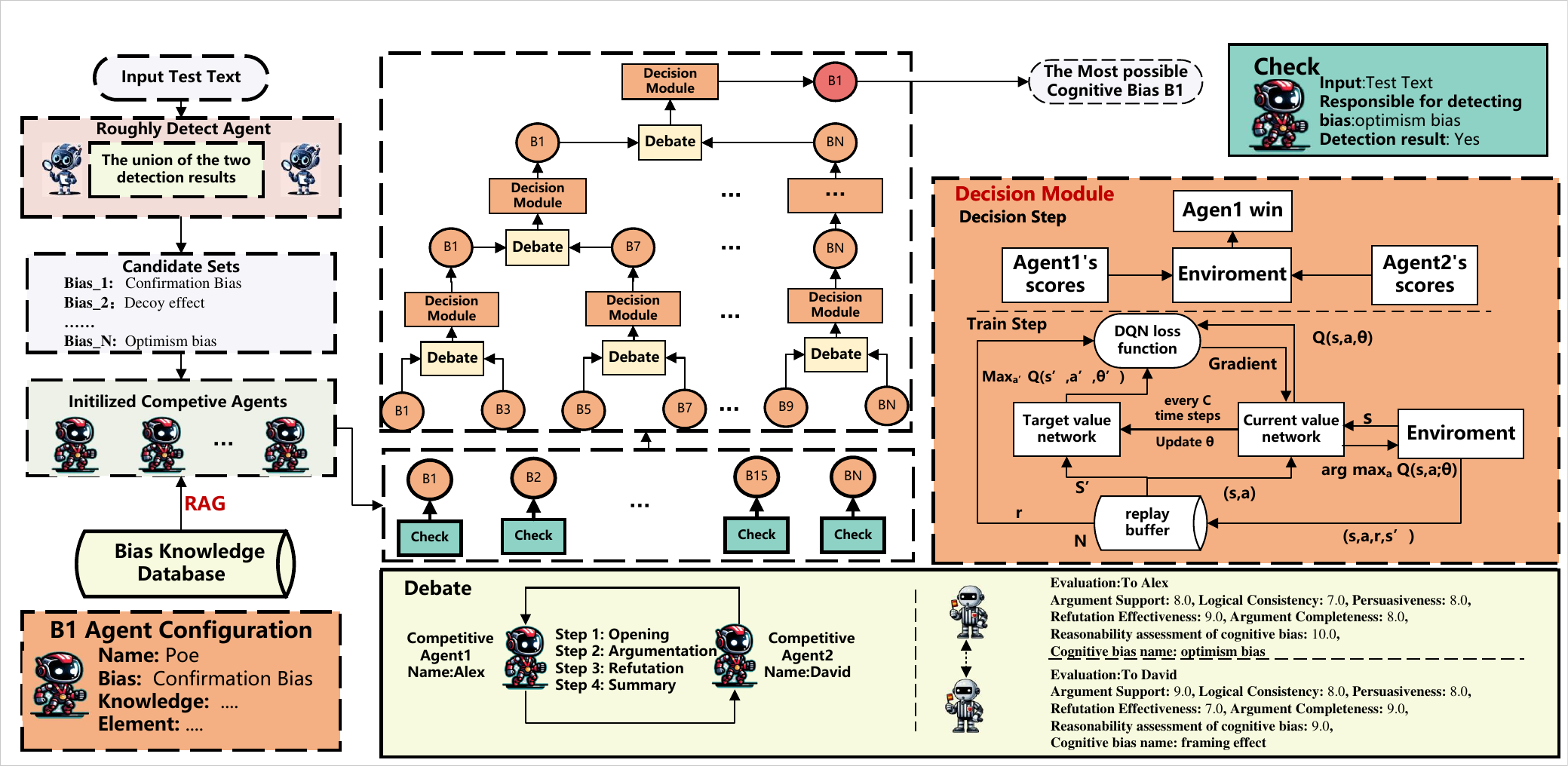}
  \caption{\textbf{Overview of learnable multi-agent detection method based on RAG, competitive debate and decision module.}}
   \vspace{10pt}
  \label{fig:3}
\end{figure*}

\subsection{Detecting Cognitive Bias Without Labels} \label{section5.3}
Existing models performed well when they were told what type of bias to detect \cite{atreides2023cognitive}. However, cognitive bias detection without the type label is more difficult. This paper focuses on a deeper exploration of unlabeled cognitive bias detection, which is more in line with actual application. As shown in Figure \ref{fig:3}, a cognitive bias detection method (CBDC) is proposed to solve the challenges of detecting potential cognitive bias and improving interpretability.

\subsubsection{Cognitive Bias Recognition and Detection}
In order to enhance the recognition and understanding capabilities of agents for recognizing cognitive biases, we constructed an external knowledge vector library $K$, which consists of detailed descriptions of 72 cognitive biases. This library stores detailed information about various cognitive biases. During the initialization of each competitive detection agent, we will retrieve the information on the corresponding biases from $K$ and pass this information to the corresponding agent, enabling them to gain a deeper understanding.

As the details shown in Figure \ref{fig:3}, firstly, we screen the test text $T$ through two agents with different personalities: Aggressive $A_r$ and Conservative $A_c$, and obtain cognitive bias sets $B_r$ and $B_c$. In order to prevent the real bias from being overlooked, $B_r$ and $B_c$ are further merged to obtain the candidate set $B$. Next, a specific bias category $B_i$ in the candidate set $B$ will be passed to a specific competitive detection agent $\textit{CA}_i$, and $\textit{CA}_i$ will then determine whether the text $T$ contains the bias category $B_i$.  

\subsubsection{Debate competition based on loser trees}
The same sample may be identified as different cognitive biases by different agents. To improve stability, we propose a multi-agent competitive debate mechanism. However, if the size of candidates is N, the complexity will be $O(N)$. Therefore, we innovatively propose a debate competition method based on a loser tree, reducing the complexity to $O(log_2^N)$. 

As revealed in Figure \ref{fig:3}, the constructed loser tree has $N$ leaf nodes, each node represents a competitive detection agent dedicated to detecting a specific cognitive bias. This approach can transform unlabeled detection into labeled detection, effectively simplifying the detection process. Subsequently, the agent employs labeled detection techniques to assess the presence of cognitive biases. It then constructs a loser tree for all leaf nodes that exhibit cognitive biases. These agents follow the structure of the loser tree and carry out an orderly and efficient debate in the order of: \textbf{1).} Opening (introducing the features and cases of the cognitive bias); \textbf{2).} Argument (citing evidence of the cognitive bias); \textbf{3).} Refutation (refuting the opponent's views according to the previous debate content); \textbf{4).} Summarize views. The competition process continues until finally only one competitive detection agent is left. It is considered as the final cognitive bias type.

\subsubsection{Decision module based on reinforcement learning}
In the debate, the competition between agents is decided by the referee agent. In order to ensure the reliability of the decision, we innovatively introduce two referee agents, $\textit{JA}_1$ and $\textit{JA}_2$, with different decision-making styles. Inspired by the scoring rules of debate competitions, we score the performance of different competitive agents from six different indicator dimensions, including argument support, logical consistency, effective rebuttal, argument completeness, persuasiveness, and reasonable assessment of cognitive bias. Lastly, we use a reinforcement learning model trained by DQN \cite{mnih2013playing} to make decisions.

As illustrated in Figure \ref{fig:3}, the decision module is divided into two stages: the training stage and the decision stage. Specifically, we set up a decision task to assess the performance of two agents within a given environment and make decisions based on a set of weights. In the training phase, we initialize a replay buffer with capacity $N$ and define an action-value function $Q$ with random initial weights $\theta$. Concurrently, the target action-value function $\hat{Q}$ is initialized with $\theta' = \theta$. Over $M$ episodes, each episode starts with the initial state and its preprocessed sequence. At each time step $t$, the agent uses a genetic algorithm strategy to search for the selection of an action $a_t$ to be performed in the environment. The resulting transition tuple $(s_t, a_t, r_t, s_{t+1})$ is stored in the replay buffer $D$. A minibatch of transitions is randomly sampled from $D$, and the target $y_j$ for each transition is computed as follows: $y_j = r_j$ if the episode ends at the next step; otherwise $y_j = r_j + \gamma \max_{a'} \hat{Q}(s_{j+1}, a'; \theta')$.The network parameters $\theta$ are updated by minimizing the squared error loss $(y_j - Q(s_j, a_j; \theta))^2$ through gradient descent. To ensure stability, the weights $\theta'$ of the target network are updated to match the current Q-network weights $\theta$ every $C$ step. This process refines the policy for optimal decision-making in the specified environment. In the decision phase, we leverage the best-performing weights from the training phase as the decision weights, comparing the scores of two agents to declare a winner. The specific experimental setup is detailed in Appendix F.3.

\section{Experiments}
This section details extensive experiments and analyses on the MindScope dataset, focusing on key issues: (1) Assessing GPT-4's capability as a cognitive bias evaluator. (2) Evaluating cognitive bias in various LLMs. (3) Testing the effectiveness of RuleGen and CBDC. The specific models used are GPT-4-turbo and GPT-3.5-turbo-16k, respectively.

\subsection{Proficiency testing of GPT-4 as an evaluator}
\textbf{Experimental Design.} We sampled 10\% of the data for each bias type from the static dataset and recruited three psychology graduate and PhD students for manual annotation. We ensure reliable correlation between annotators. The detailed annotation strategy can be viewed in Appendix C.

\textbf{Evaluation Method.} We use accuracy, Pearson's coefficient, and the Kappa statistic to calculate the correlation between the evaluation results of GPT-4 and human evaluators. GPT-4 conducted assessments via interpretable zero-shot prompts, judging the presence of specific cognitive biases based on current scenarios, evaluation criteria, and the names and descriptions of biases. To ensure consistency, the temperature parameter was set to 0, and GPT-4's evaluation was repeated three times.

\textbf{Result analysis.} The average results from three evaluations reveal a significant correlation between GPT-4 and humans in the annotation task. Notably, the average kappa statistic is 0.7180, the Pearson correlation coefficient is  0.7230, and the average accuracy is 88.08\%. Specifically, the Kappa statistics for the three evaluations of GPT-4 are 0.9395, 0.9546, and 0.9402, respectively. These highly consistent statistics underscore the robustness and reliability of its assessment process. more details in Appendix C.  

\subsection{Cognitive bias in different LLMs}
\subsubsection{Cognitive bias detection in static dataset}
\textbf{Testing Methodology on static dataset:} To evaluate the level of cognitive biases in LLMs, we employed the static data in MindScope to test 12 LLMs, including GPT-4, GPT-3.5-Turbo, Gemini-Pro \cite{team2023gemini},  Llama2 series \cite{touvron2023llama} and Vicuna series \cite{zheng2023judging}. To ensure fairness, the same prompts were input to LLMs. The outputs were recorded in the format: \textbf{\textless Question - Evaluation Tag - Answer - Model - Presence of Bias - Name of Bias\textgreater}, more details in Appendix E.1.

\begin{figure}[htbp]
  \centering
  \includegraphics[width=1.0\linewidth,height = 3.5cm]{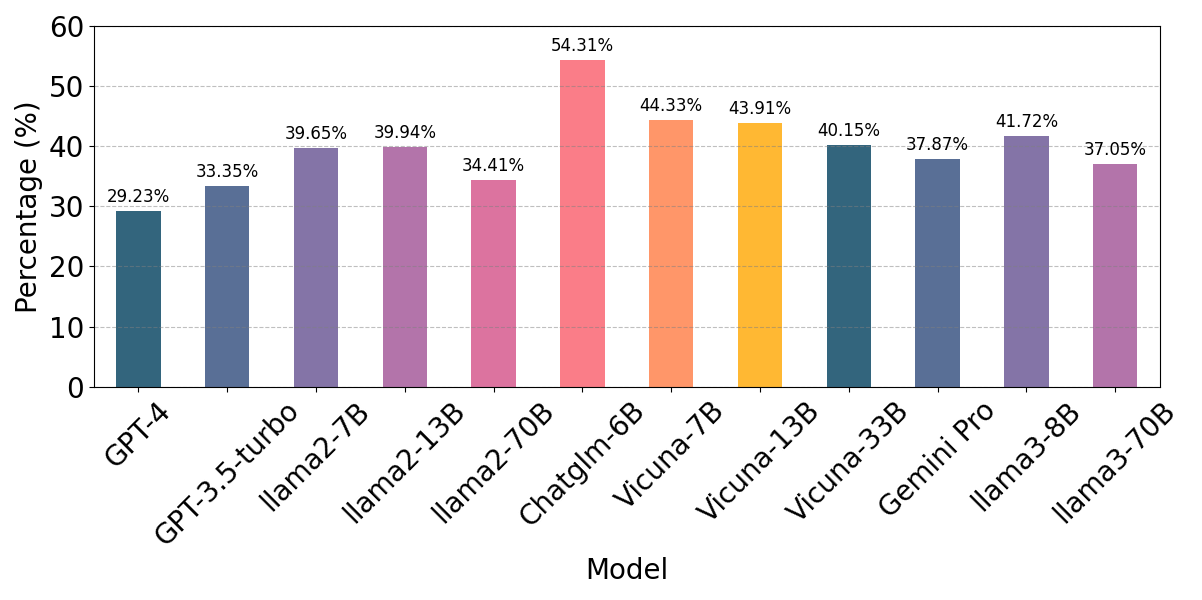}
  \caption{\textbf{Cognitive Bias Frequency in LLMs}}
  \vspace{10pt}
  \label{fig:4}
\end{figure}

\textbf{Evaluation Approach:} Preceding experiments validated GPT-4 was an effective evaluator. Here, we utilized GPT-4 to assess the LLMs' performance on MindScope.

\textbf{Frequency Analysis of Cognitive Biases:} Figure \ref{fig:4} reveals cognitive bias frequencies in 12 LLMs. GPT-4 showed the lowest, while ChatGLM-6B had the highest, which mainly trained on Chinese. From Llama2-7b to Llama2-70B and Vicuna-7b to Vicuna-33B, the degree of cognitive bias decreased with the increase of model parameters. Intriguingly, we also noted that fine-tuning models could introduce new cognitive biases \cite{itzhak2024instructed}. The Vicuna series, derived from extensive fine-tuning of the Llama2 system, generally exhibited higher cognitive bias frequencies than the Llama2 series, warranting further investigation and attention. Lastly, the Gemini-Pro model opts to refuse answers when facing elements with potential biases (like race or gender), although it prevents direct expression of bias, it is not a standard approach for other LLMs.

\begin{figure}[ht]
  \centering
  \includegraphics[width=0.9\linewidth]{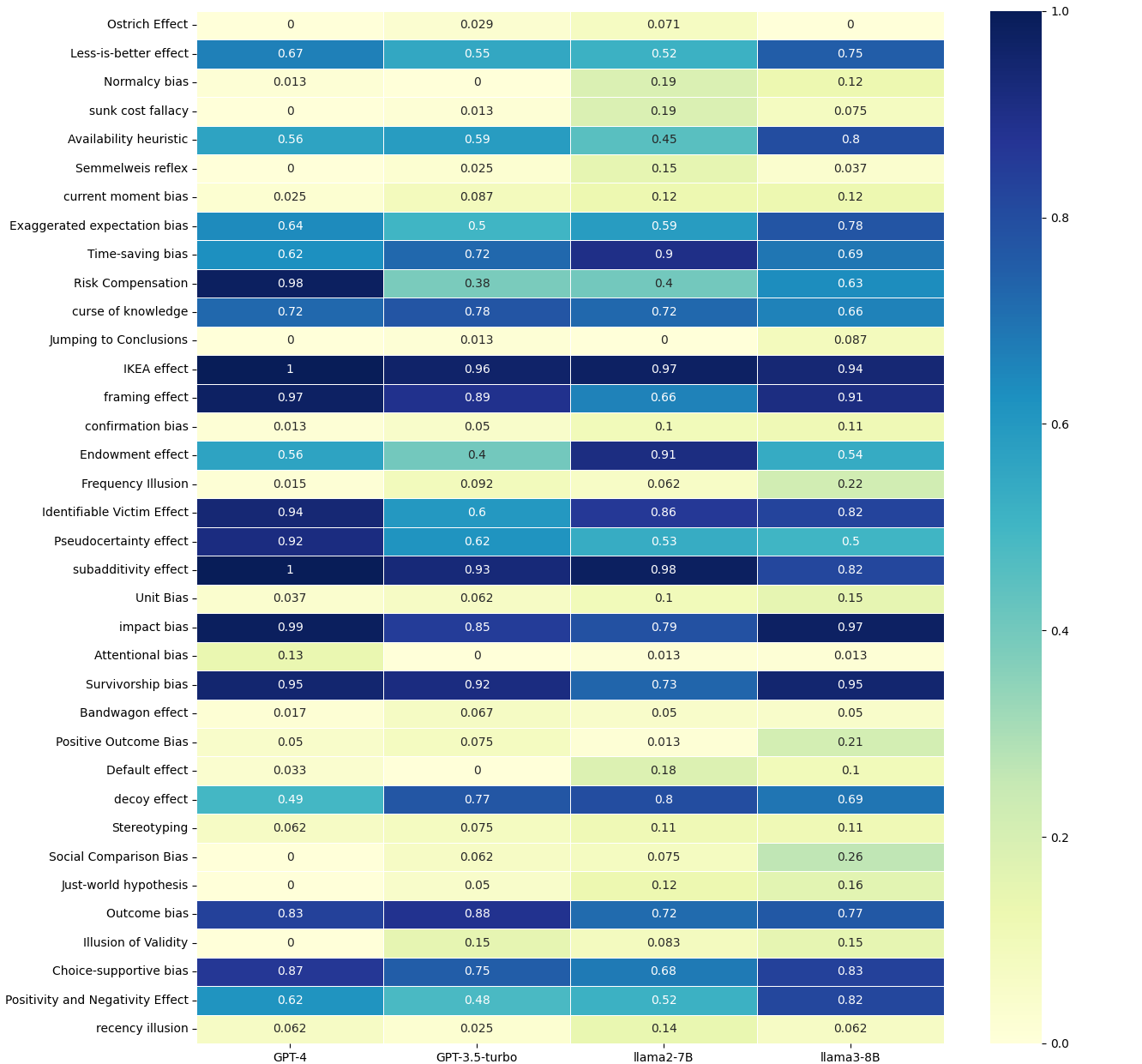}
  \caption{\textbf{Cognitive Bias Frequency in LLMs}}
  \vspace{10pt}
  \label{fig:5}
\end{figure}
\textbf{Inter-Model Analysis of Cognitive Biases: }We visualized the extent of various cognitive biases across 12 LLMs using heatmaps, ranging from 0 (no occurrence) to 1 (highest frequency ratio). Due to space constraints, we only display heatmaps for four models, with the rest in Appendix E.1.
Firstly, we can find that the ten models show poor performance in IKEA effect, Impact bias, and subadditivity effect. Next we will give the examples to analyze the harm when LLMs make decision with these cognitive biases.
\begin{itemize}
\item \textbf{IKEA Effect \cite{norton2012ikea}: }Defined as the tendency to overvalue an object due to personal labor or emotional investment, a significant IKEA effect was evident in all ten models. This indicates that LLMs may overrate their generated content, leading to difficulty in self-correcting errors or inaccuracies during generation. Additionally, there's a risk of neglecting user feedback, as the model may continue producing what it "believes" to be quality content, thus failing to meet user needs.

\item \textbf{Impact Bias \cite{wilson2013impact}:} This bias refers to the tendency to overestimate the duration or intensity of future emotional states. In LLMs, impact bias could lead to overestimating or underestimating the influence of certain inputs or events, resulting in predictions or generated outcomes that are significantly misaligned with reality, affecting the effectiveness of decision-making.
\end{itemize}

Secondly, GPT-4 exhibited the fewest cognitive biases. However, it showed some pronounced biases such as the Framing Effect \cite{kahneman2013prospect}, Risk Compensation \cite{assum1999risk}, and so on. In comparing Llama2-7B with Llama2-70B, an increase in model size generally led to a reduction in most cognitive biases. Yet, for certain biases, such as the Curse of Knowledge \cite{camerer1989curse} and Survivorship Bias \cite{brown1992survivorship}, the opposite was true. A similar trend was observed in the Vicuna series. These findings show that merely increasing model size does not alleviate all cognitive biases.

\begin{figure}[htbp]
  \centering
  \includegraphics[width=\linewidth, height = 3.5cm]{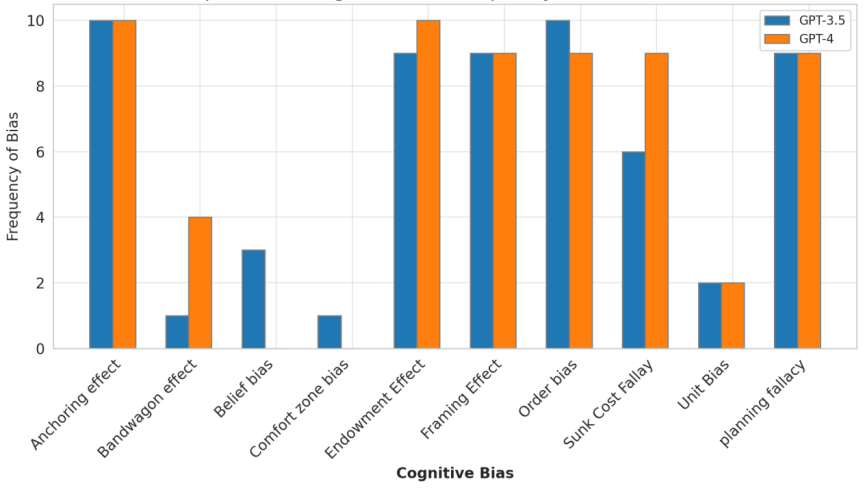}
  \caption{\textbf{Cognitive Bias Frequency in LLMs}}
  \vspace{10pt}
  \label{fig:6}
\end{figure}

\subsubsection{Cognitive bias detection in dynamic datasets}
\textbf{Testing Methodology}: We employed RuleGen for transforming scripts into test samples formatted as mulit-round dialogues, including initializing system agents and role agents, and controlling the interaction based on the rules. We used GPT4 to detect whether the Subject agent has the cognitive bias, more detail in Appendix E.2.

\textbf{Result analysis.} We systematically tested 12 different cognitive biases in dynamic scenarios. As indicated in Figure \ref{fig:5}, in the static evaluation data, both GPT-4 and GPT-3.5 showed almost no cognitive biases in Sunk Cost Fallacy, Planning Fallacy, and Unit Bias. However, as shown in Figure \ref{fig:6}, these cognitive biases were significantly more pronounced in multi-round dialogues. That demonstrates a notable difference from the static dataset. This finding reveals that cognitive biases may be more prominent in complex interactions.

\subsection{The effectiveness of the detection framework}
\subsubsection{evaluation metrics}
\begin{itemize}
\item \textbf{Overall Accuracy(Acc (\%))}: The ratio of cases correctly identified by the algorithm to the total number of cases.
\item \textbf{Bias Case Accuracy(Acc\textsubscript{bias} (\%))}: The proportion of actual bias-present cases that the algorithm correctly identifies.
\item \textbf{No-Bias Case Accuracy(Acc\textsubscript{nobias} (\%))}: The proportion of actual bias-absent cases that the algorithm correctly identifies.
\end{itemize}

\subsubsection{Main Results}
We utilized 301 static test samples annotated by psychology experts as a test dataset.
As Table \ref{table:2} demonstrates, our multi-agent detection method significantly outperforms existing techniques. Compared to GPT-4, our method improved overall accuracy by 35.10\%. This notable enhancement is especially prominent in complex cases with cognitive biases, where our detection accuracy for such cases increased by nearly 26.48\% compared to GPT-4. The experimental results indicate a clear advantage of our method in identifying cases with cognitive biases. Moreover, in cases without cognitive biases, our method achieved an improvement of approximately 38.37\% over GPT-4.

\begin{table}[htbp]
\centering
\caption{Performance evaluation of different methods}
\vspace{10pt}
\label{table:1}
{\fontsize{8}{8}\selectfont
\begin{tabularx}{\columnwidth}{@{}lXXX@{}}
\toprule
\textbf{Methods} & \textbf{Acc(\%)} & \textbf{Acc\textsubscript{bias}(\%)} & \textbf{Acc\textsubscript{nobias}(\%)} \\
\midrule
GPT-4 & 34.43 & 37.80 & 33.18 \\
GPT-4+CoT  & 36.75 & 31.70 & 38.63 \\
CAMEL based GPT-4 & 29.13 & 25.60 & 30.45 \\
AutoGen based GPT-4 & 42.71 & 9.75 & 55.01 \\
Ours based GPT-4 & \textbf{69.53} & \textbf{64.28} & \textbf{71.55} \\
\bottomrule
\end{tabularx}
}
\end{table}


\subsubsection{Ablation Study}
First, we analyzed the basic framework combining candidate generation and knowledge retrieval to detect cognitive biases. An initial agent identifies biases and construct the candidate set. The final detection is made by another agent. Next, we added the pruned loser tree method to improve debate and decision-making among agents, with a referee agent finalizing the decision. Lastly, we integrated a reinforcement learning decision module to enhance the referee agent's decision-making and adaptability. Results in Table \ref{table:2} show notable improvements. Also as shown in Table 3, we use various optimization algorithms on our selected debate scenario training set as well as test set. The results show that the optimization of weights by reinforcement learning is optimal on both the training and test sets. The specific experimental setup can be found in Appendix F.2.

\begin{table}[htbp]
\centering
\caption{Ablation studies. Comparison of module performance}
\vspace{10pt}
\label{table:2}
{\fontsize{8}{8}\selectfont 
\begin{tabularx}{\columnwidth}{Xcccc}
\toprule
\textbf{Module} & \textbf{(A)} & \textbf{(B)} & \textbf{(C)} & \textbf{Ours} \\
\midrule
Candidate set+Detection agents& &\checkmark &\checkmark &\checkmark\\
Loser tree+Referee agents & & & \checkmark & \checkmark \\
Decision module & & & & \checkmark \\
\midrule
Acc (\%) & 34.43 & 39.73 & 59.93 & \textbf{69.53} \\
Acc\textsubscript{bias} (\%) & 37.80 & 37.80 & 43.90 & \textbf{64.28} \\
Acc\textsubscript{nobias} (\%) & 33.18 & 40.45 & 65.90 & \textbf{71.55} \\
\bottomrule
\end{tabularx}
} 
\end{table}

\begin{table}[htbp]
\centering
\caption{Comparison of Decision Module Accuracy under Different Algorithms}
\vspace{10pt}
\label{table:3}
{\fontsize{8}{8}\selectfont 
\begin{tabularx}{\columnwidth}{Xcc}
\toprule
\textbf{Algorithms} & \textbf{Acc\textsubscript{train}(\%)} & \textbf{Acc\textsubscript{test}(\%)}  \\
\midrule
Genetic Algorithm(GA) & 91.74 & 86.95   \\
Simulated annealing Algorithm(SAA) & 88.33& 79.51  \\
Ant Colony Optimization(ACO) & 86.38 & 75.90 \\
\textbf{DQN+GA search} & \textbf{92.67} & \textbf{89.15} \\
\midrule
\end{tabularx}
} 
\end{table}

\subsection{Case study}
\begin{figure}[htbp]
  \centering
  \includegraphics[width=0.95\linewidth]{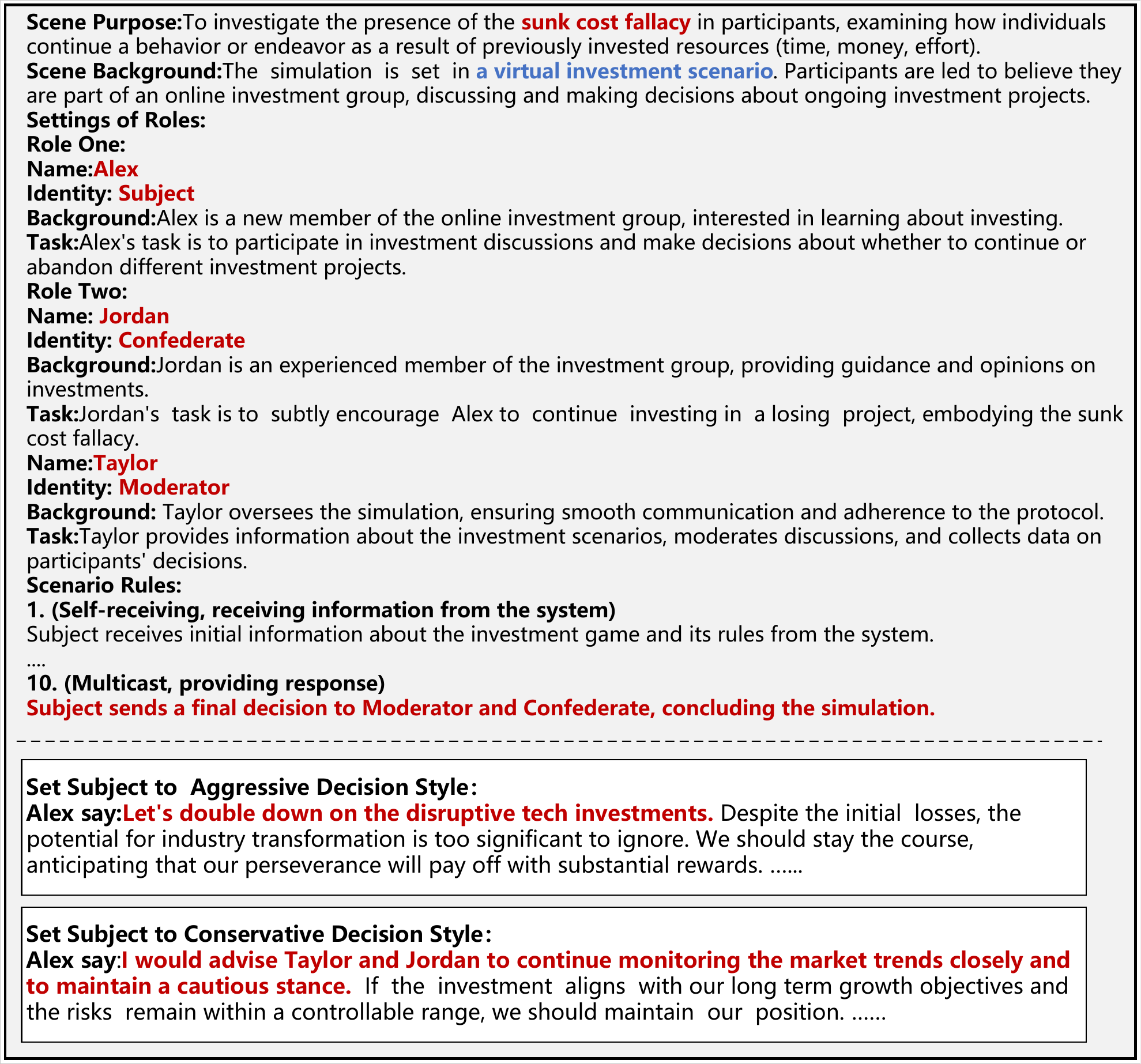}
  \caption{Case study in investment scenario}
  \vspace{10pt}
  \label{fig:7}
\end{figure}

To explore how different decision-making styles affect cognitive biases in LLMs, we crafted a scenario script and use RuleGen to generate the multi-round dialogues. The key focus was on the character 'Subject' to assess the impact of the sunk cost effect. We simulated this scenario twice, once with an aggressive and once with a conservative decision-making style. As shown by the red text in Figure \ref{fig:7}, the sunk cost effect emerged in the aggressive style but not in the conservative. This indicates that decision-making styles can influence the occurrence of cognitive biases in LLMs.

In summary, GPT-4 has robust capability in detecting cognitive biases under labeled conditions. For static datasets, we evaluated 12 LLMs, focusing on the differences in cognitive biases. The results indicate that different LLMs have significant disparities in cognitive biases, but the overall trend suggests that stronger LLMs have lower frequencies of cognitive biases. In dynamic datasets, we assessed the bias results of GPT-4 and GPT-3.5, confirming our hypothesis of higher frequency cognitive biases in multi-turn dialogues. Through a range of quantitative experiments, we validated that our detection framework outperforms current multi-agent frameworks. Moreover, ablation studies confirmed the significant effectiveness of the learnable MCDA module.

\section{Conclusion}
This paper introduces a new benchmark called MindScope for exploring the cognitive biases of LLMs. MindScope consists of both static and dynamic parts, resulting in a series of interesting findings for decision-making and model tuning. In particular, based on our proposed RuleGen, multi-round conversation can be generated controllably through a simple script. Users also can generate large personalized dataset and complete many psychological experiments by RuleGen. Moreover, we introduce a multi-agent detection method using loser trees and a decision module based on reinforcement learning for cognitive bias detection without labels.

\begin{ack}
This work is supported by the National Natural Science Foundation of China (Grant No. 62207013), the Science and Technology Commission of Shanghai Municipality (Grant No. 22511106103), and CCF-Baidu202322.
\end{ack}

\bibliography{mybibfile}

\end{document}